# Two new feature selection methods based on learn-heuristic techniques for breast cancer prediction: A comprehensive analysis

Kamyab Karimi[a], Ali Ghodratnama [a,*], Reza Tavakkoli-Moghaddam[b]

[a] *Department of Industrial Engineering, Faculty of Engineering, Kharazmi University, Iran, Tehran*
[b] *School of Industrial Engineering, College of Engineering, University of Tehran, Iran, Tehran*

## Abstract

In recent decades, breast cancer has become one of the leading causes of mortality among women. This disease is not preventable because of its unknown causes; however, its early diagnosis increases patients' recovery chances. Machine learning (ML) can be utilized to improve treatment outcomes in healthcare operations while diminishing costs and time. In this research, we suggest two novel feature selection (FS) methods based upon an imperialist competitive algorithm (ICA) and a bat algorithm (BA) and their combination with ML algorithms. This study aims to enhance diagnostic models' efficiency and present a comprehensive analysis to help clinical physicians make much more precise and reliable decisions than before. *K*-nearest neighbors (KNN), support vector machine (SVM), decision tree (DT), Naive Bayes, AdaBoost (AB), linear discriminant analysis (LDA), random forest (RF), logistic regression (LR), and artificial neural network (ANN) are some of the methods employed. Sensitivity, accuracy, precision, mean absolute error F-score, root mean square error, Kappa, relative absolute error calculated the performance of the methods. This paper applied a distinctive integration of evaluation measures and ML algorithms using the wrapper feature selection based on ICA (WFSIC) and BA (WFSB) separatly. We compared two proposed approaches for the performance of the classifiers. Also, we compared our best diagnostic model with previous works reported in the literature survey. Experimentations were performed on the Wisconsin diagnostic breast cancer (WDBC) dataset. Results reveal that the proposed framework that uses the BA with an accuracy of 99.12%, surpasses the framework using the ICA and most previous works. Additionally, the RF classifier in the approach of FS based on BA emerges as the best model and outperform others regarding its criteria. Besides, the results illustrate the role of our techniques in reducing the dataset dimensions up to 90% and increasing the performance of diagnostic models by over 99%. Moreover, the result demonstrates that there are more critical features than the optimum dataset obtained by proposed FS approaches that have been selected by most ML models, including the standard error of area, concavity, smoothness, perimeter, the worst of texture, compactness, radius, symmetry, smoothness, concavity, and the mean of concave points, fractal dimension, compactness, concavity that can remarkably affect the efficiency of breast cancer prediction. This study illustrates the role of our approaches for enhancing treatment outcomes in healthcare operations.

---
[*] Corresponding author. Tel.: +98 21 88830891.
*E-mail address:* ghodratnama@khu.ac.ir (A. Ghodratnama).

## 1. Introduction

Compared to other types of cancer, breast cancer is considered one of the significant cancer types, the second most prevalent cause of death among women[1] (Chaurasia and Pal, 2014). In 2014, an estimated 232714 new breast cancer cases were diagnosed in women, with 297800 female patients dying (Mandal, 2017). In 2020, there will be 2.3 million diagnoses and 685000 deaths worldwide. Breast cancer had been diagnosed in 7.8 million women alive in the previous five years by the end of 2020, making it the world's most common cancer[2]. The American Cancer Society (ACS) reports that 99 percent of people who receive treatment for breast cancer in the earliest stages live for 5 years or longer after diagnosis[3].

The methods for cancer diagnosis are primarily dependent on the expertise of physicians whose medical imaging is used to assist in this process and diagnose specific symptoms generally occurring in later stages of cancer. This disease is treatable in the early stages when symptoms and clinical signs appear in advanced stages (Li and Chen, 2018). Therefore, diagnosing in the early stages can save the lives of many patients. Also, due to human and healthcare operations errors, various treatments that experts and doctors accomplish might not be effective enough, and timeliness is also costly[4].

Misdiagnosis leads to incorrect treatments, which causes the patients to lose the best time for treatment, which will prompt drastic results (Reddy et al., 2022). As a result, selecting a framework to predict the nature of the breast tumor and categorizing it as benign or malignant is of utmost importance.

Machine learning (ML), which is one of the most widely used branches of artificial intelligence (AI), engages with the development and investigation of methods and algorithms that provide computers and systems with the ability to train and learn. ML greatly assists in saving operational costs and improving the speed of the data analysis (Russel and Norvig, 2013). For example, in recent years, with the increase of community knowledge and AI development, especially in data mining, many studies have been conducted regarding the early detection of breast cancer (Mate and Somai, 2021). On the one hand, classification models exploited in data-mining methods determine the type of tumor and the accuracy of diagnosis. However, on the other hand, vast amounts of information and features in a dataset consume much time and create high computational costs for ML algorithms to process this amount of data (Tohka and van Gils, 2021). Hence, reducing the dimensions of the dataset to its more critical information content is of particular importance.

---

[1] https://www.statista.com/statistics/1031260/cancer-death-rate-worldwide-by-type

[2] https://www.who.int/news-room/fact-sheets/detail/breast-cancer

[3] https://www.healthline.com/health/breast-cancer/breast-cancer-cure#survival-rates

[4] https://www.who.int/news-room/fact-sheets/detail/patient-safety

Feature selection (FS) plays an essential role in ML algorithms. It aims to remove the irrelevant or redundant features of the data to achieve better classification performance (Qasim, O. S., & Algamal, 2020). As a result, FS methods improve breast cancer prediction and diagnosis accuracy in less time and during the early stages of cancer by selecting a set of practical features among the features available in the primary dataset (Zebari et al., 2020).

The common classification problems are based on the plan of classifiers dealing with the type of operational parameters chosen. Fine-tuning the parameters needs skill. Instead, metaheuristic algorithms simplify planning non-parametric classifiers, classifying the data in a straight line. Meta-heuristic algorithms make us close to the best solution even in a complex search space. In addition, they can also deviate from the local optimum solution (Hosseini et al., 2014). Consequently, these two mentioned merits of meta-heuristics algorithms make them capable of generating exact and robust solutions in a shorter time than medical ways.

Related to the bat algorithm (BA) and the imperialist competitive algorithm (ICA) for FS, there are substantial grounds to consider them indispensable. To begin with, they are based on a systematic mathematical computation, making it easy for researchers to look into convergence and robustness. They are also effective alternatives for optimizing functions with many decision variables (Hosseini and Al Khaled, 2014). Besides, they do not require complicated mathematical calculations and operators for FS. Accordingly, they have been effectively practical to solve numerous optimization problems like facility layout (Wang et al., 2014), assembly line balancing (Lei et al., 2019), scheduling (Rashid et al., 2021; Ahmadi et al., 2013), artificial neural network optimization (Bahmani et al., 2021; Lu et al., 2020), and healthcare (Hosseini et al., 2014). Furthermore, they are effective global optimization methods and have strong local searchability with a flexible structure (Hosseini and Al Khaled, 2014). It is not essential to increase local search capability, such as a genetic algorithm (GA). These features on an imperialist competitive algorithm (ICA) and a bat algorithm (BA) stimulate us to apply them for breast cancer prediction.

In different situations, different criteria are suitable, and it is not always apparent which criterion to employ. Another issue is that machine learning algorithms that perform well on one criterion may not perform well (Caruana and Niculescu-Mizil, 2004). In other words, one of the crucial predicaments that previous studies have is that most of them have employed just a small number of measures (Mandal, 2017; Rao et al., 2019; Mahendru and Agarwal, 2019; Liu et al., 2019; Aslan et al., 2018). As a result, we use nine assessment criteria: accuracy, sensitivity, specificity, precision, F-measure, Kappa statistic, root mean square error (RMSE), mean absolute error (MAE), and relative absolute error (RAE). Furthermore, it should be noted that each ML algorithm has is that they have its advantages and disadvantages and is executed differently for a dataset. Therefore, we experiment with nine ML algorithms to ensure that our best model is more reliable, while previous works have considered fewer ML algorithms that inevitably have less validity (Abdar et al., 2020; Liu et al., 2019; Wang et al., 2018; Mahendru and Agarwal, 2019).

This study proposes a framework that includes the wrapper feature selection based on imperialist competitive (WFSIC) and bat (WFSB) separately to integrate with ML algorithms. This approach has been applied to perform tumor classification and breast cancer prediction in the early stages with less time and cost and much higher accuracy. To comprehend our paper objective better, we similarly decide to reply to some research queries, are as follows:

- What are the most critical features that significantly affect breast cancer?
- How many are there the best features for each ML model?

Motivated by these questions, this study aims to find the best model among the classification models for breast cancer diagnosis. We introduce a framework to perform this task with higher accuracy and shorter time and cost consumption. Our framework integrates the WFS method based on the proposed meta-heuristic algorithms with ML algorithms. The performance of diagnostic models is measured using evaluation criteria. The best model is then obtained by comparing and analyzing their performances with and without WFSIC and WFSB. Finally, we compare the efficiency of the two proposed approaches with previous works that have used FS methods in terms of criteria used in this study. Then, we recognize which classification model and algorithms accomplish better than the others do for breast cancer prediction. As a result, the following are the significant contributions of our work:

- Considering the advantages of the ICA, BA, and WFS, we propose the two new WFSIC and WFSB approaches integrated with ML algorithms, removing redundant and irrelevant features from the dataset, improving the classification performance, and reducing computational cost and time.
- Considering that each ML method has advantages and disadvantages. It performs differently depending on the dataset. We investigate nine ML algorithms and nine criteria together, which have never been used so far. It makes our study more reliable and comprehensive than other works.

The rest of this work is organized as follows: The second section goes over the prior efforts. The methodology and instruments employed in this investigation are proposed in Section 3. Section 4 contains the comprehensive analysis and outcomes. Finally, Sections 5 and 6 bring this study to a close and provide recommendations for future research.

## 2. Literature survey

Extensive studies have investigated and diagnosed breast cancer using different data mining and ML methods. This section reviews two subjects, including diagnosis without and with FS.

## 2.1. Diagnosis without FS methods

Some studies performed in this area have merely compared ML methods. For instance, various data mining methods and classification algorithms have been used to compare methods' performance and not to improve them. Support vector machine (SVM) and artificial neural network (ANN) approaches are examined as ML methods for classification in the WEKA tool by Bayrak et al. (2019). They have compared the results obtained from these methods using the parameters of the Accuracy, Precision, recall, and receiver operating characteristic (ROC) diagram. The SVM has shown the best performance in the accuracy of 96,9957%. Alshammari and Mezher (2020) provided Using the WEKA tool, and a comparative analysis was carried out. Lazy IBK (instance-based learning with parameter K), Naive Bayes (NB), logistic regression (LR), DT, lazy locally weighted learner, lazy k-star, decision stump, rules zeroR, RF, and random trees (RT) were among the techniques shown. The accuracy ratios for all studies were subtracted from the final evaluation of grouping procedures, revealing high proportions ranging from 72 to 98 percent. Islam et al. (2020) offered a qualified study of five ML methods: SVM, K-nearest neighbors (KNN), RF, ANN, and LR. The negative predictive value, false-negative rate, Accuracy, Sensitivity, Specificity, Precision, F1-score, false-positive rate, were used for model evaluation, and the ANN achieved the highest Accuracy with 98.57 percent. Abdel-Ilah and Šahinbegovi (2017) implemented a feed-forward backpropagation network (FFBPN) to classify breast cancer. The research aimed to plan an ANN with an acceptable and high level of accuracy by choosing the number of hidden layers. The results showed that the best system design had three hidden layers and 21 neurons in a hidden layer. Salehi et al. (2020) introduced two ML methods based on the Multi-Layer Perceptron (MLP) learner machine, involving MLP, stacked generalization, and a combination of MLP-experts. It was performed on the Investigation, End Results and Epidemiology (SEER) database to investigate their ability for survivability estimate of breast cancer patients. The assessment of the predictors discovered an accuracy of 83.86% and 84.32% by the combination of MLP- stacked generalization and MLP-experts methods.

An approach was proposed by Mojrian et al. (2020). ELM-RBF was a multilayer fuzzy expert system that detects breast cancer using an extreme learning machine (ELM) grouping model and a radial basis function (RBF) kernel. With an accuracy of 95.69 percent, the planned multilayer fuzzy expert system of ELM-RBF surpassed the SVM model in terms of sensitivity, accuracy, validation, precision, false-positive rate, specificity, and false-negative rate. Abdar et al. (2020) presented a two-layer nested cooperative method where voting and stacking were adopted for grouper mixtures in ensemble approaches for breast cancer diagnosis and compared the developed approach with other single groupers conducted on the WDBC dataset. The associated results revealed that their models outdid diverse single classifiers. Both of them achieved an accuracy of 98.07%. Mangukiya et al., (2022) presented data Visualization and performance comparisons between different ML algorithms: SVM, DT, NB, KNN, AB, XGboost and RF conducted on the WDBC dataset. They concluded that XGboost is the most accurate algorithm for best accurate result for detection of breast cancer with the efficiency of 98.24%. Kumar et al., (2021) proposed a comparative analysis of ML techniques. The techniques used for detecting and diagnosing the breast cancer are SVM, RF, and KNN. All experiments had been conducted on the WDBC dataset. Their results showed that KNN had achieved the highest accuracy (97.32%) compared to SVM and RF.

## 2.2. Diagnosis with FS methods

FS has been a hot topic of study in recent years. A detailed literature study on feature selection strategies was undertaken in several kinds of research. On the Wisconsin prognosis breast cancer (WPBC) dataset, Sakri et al. (2018) focused on improving the accuracy value utilizing an FS approach called particle swarm optimization (PSO), as well as reduced error pruning (REP), tree KNN, and NB. The Saudi Arabian women's breast cancer issue was held from their endeavor standpoint. According to their findings, women over the age of 46 are the most common victims of this deadly illness. Salma and Doreswamy (2020) proposed a hybrid diagnostic model that combines the feed-forward neural network (FNN), the Bayesian algorithm (BA), and the gravitational search algorithm (GSA). According to the research, the hybrid BATGSA model outperforms its counterparts, generating a high accuracy of 94.28 percent for the WDBC dataset. Sangaiah and Kumar (2019) presented a hybrid FS algorithm regarding ReliefF and Entropy GA for breast cancer diagnosis on the WBC dataset. The accuracy obtained from the model was 85.89%.

Chaurasia and Pal (2020) adopted a novel FS statistical approach to increase forecast accuracy. They compared classification and regression tree (CART), SVM, NB, KNN, LR, and MLP, all machine learning techniques. LR performed better than the other classifiers on the WDBC dataset, with an accuracy of 95.1739 percent. The FS was given an information gain (IG) based on simulated annealing (SA) and genetic algorithm (GA) wrapper (IGSAGAW) by Liu et al. (2019). The IG is often used in training decision tree (DT) models by comparing the entropy of the dataset before and after a transformation. After removing the top-the-best feature, they used the cost-sensitive support vector machine (CSSVM) learning technique to classify features linked to the IG algorithm. According to the findings acquired from the WDBC dataset, the CSSVM had a maximum accuracy of 95.8%. Oladele et al. (2021) used Ant Colony Optimization (ACO) and PSO to offer meta-heuristic optimization techniques for breast cancer detection. The diagnosis model was assessed using the Wisconsin Original Breast Cancer (WOBC) dataset's Accuracy, precision, recall, and F1-measure. The maximum accuracy was found to be 97.1388 percent in an experimental assessment. Fan and Chaovalitwongse (2010) suggested a novel feature selection optimization approach for medical data categorization. Support Feature Machine was the name given to this framework (SFM). They were tested on five real medical datasets, including epilepsy, breast cancer, heart illness, diabetes, and liver problems. The results obtained from the WDBC dataset revealed a sensitivity of 96.22% and a specificity of 99.19%. Rao et al. (2019) proposed a new feature selection technique based on a bee colony and a gradient boosting decision tree to solve feature efficiency and informative quality issues. Their trials employed two breast cancer datasets and six datasets from a public data source. On the WDBC dataset, the results demonstrated an accuracy of 97.18%. Mate and Somai (2021) introduced bayesian optimization technique along with hyper parameter tuning and feature selection techniques like pearson's coefficient, chi square test ,logistic regression, random forest. They obtained an accuracy of 96.2%, with Extra tree classifier algorithm on the WDBC dataset. Rajendran et al., (2022) presented a hybrid optimization algorithm that combines the grasshopper optimization algorithm and the crow search algorithm for feature selection and classification of the breast mass with multilayer perceptron. Their proposed model achieved an accuracy of 97.1%, a sensitivity of 98%, and a specificity of 95.4% for the mammographic image analysis society dataset.

### 2.3. Research gap

As revealed in Table 1, several implications can be drawn from the literature review of studies related to diverse feature selection and ML algorithms. First, most papers used just a few measures like Accuracy[1] (Liu et al., 2019; Rao et al., 2019; Mahendru and Agarwal, 2019) or other common criteria such as sensitivity and specificity (Bayrak et al., 2019; Salma and Doreswamy, 2020; Wang et al., 2018) that could not be comprehensive for making more precise decisions, since there was not always a specific criterion to use, and multiple criteria were needed for multiple settings. Likewise, ML algorithms that operate well on one criterion may not operate well on other criteria (Sokolova and Lapalme, 2009; Caruana and Niculescu-Mizil, 2004). Thus, in addition to the mentioned criteria, we exert other applicable measures such as precision, F-score, Kappa statistic, RMSE, RAE, and MAE to raise the validity of the research. Second, the use of evolutionary algorithms as search techniques in the wrapper approach upturned the overall efficiency of the ML algorithms. As a result, we offer two novel feature selection approaches that combine the ICA and BA with ML algorithms. The WFSIC and WFSB aim to eliminate irrelevant data with less predictive output and select more essential features for increasing accuracy and performance of diagnosis at early stages, since the selection of more important data from the dataset results in saving the costs of collecting medical data and reducing the time of diagnosis and patients' wait for treatment. Third, previous works have utilized less than five classifiers to diagnose breast cancer that can have less comprehensively and reliability.

In contrast, all nine possible ML algorithms for classification problems have not been exploited to predict breast cancer (Bayrak et al., 2019; Abdar et al., 2020; Salma and Doreswamy, 2020; Liu et al., 2019; Rao et al., 2019; Wang et al., 2018; Mahendru and Agarwal, 2019). Using nine ML algorithms aims to enhance decision-making reliability and present a comprehensive analysis for clinical physicians. These classifiers are NB, SVM, KNN, LR, DT, ANN, linear discriminant analysis (LDA), and two ensemble learning algorithms, named AB and RF. Table 1 classifies the related literature review and the status of our contribution.

---

[1] https://www.who.int/news-room/fact-sheets/detail/breast-cancer

**Table 1**

Comparison between previous research and this study

| Authors | Year published | Method | Evaluation measure | | | | | | | | | Machine learning algorithm | | | | | | | | |
|---|---|---|---|---|---|---|---|---|---|---|---|---|---|---|---|---|---|---|---|---|
| | | | Accuracy | Sensitivity | Specificity | Precision | F-score | Kappa | MAE | RMSE | RAE | ANN | SVM | NB | KNN | LR | DT | RF | AB | LDA |
| Islam et al. | 2020 | - | ✓ | ✓ | ✓ | ✓ | ✓ | | | | | ✓ | ✓ | | ✓ | ✓ | | ✓ | | |
| Abdar et al. | 2020 | SV-Naïve Bayes | ✓ | ✓ | | ✓ | ✓ | | | | | | ✓ | ✓ | | | | | | |
| Bayrak et al. | 2019 | - | ✓ | ✓ | | ✓ | | | | | | ✓ | ✓ | | | | | | | |
| Liu et al. | 2019 | IGSAGAW | ✓ | | | | | | | | | ✓ | ✓ | | | | | | | |
| Mojrian Et al. | 2020 | RF-EGA | ✓ | ✓ | ✓ | ✓ | ✓ | | | | | | ✓ | ✓ | ✓ | | ✓ | ✓ | | |
| Salma and Doreswamy | 2020 | Hybrid BATGSA | ✓ | ✓ | ✓ | ✓ | | | | | | ✓ | | | | | | | | |
| Wang Et al. | 2018 | SVM-based ensemble | ✓ | ✓ | ✓ | | | | | | | | ✓ | | | | | | | ✓ |
| Mandal | 2017 | PCC | ✓ | | | | | | | | | | | | ✓ | | ✓ | ✓ | | |
| Aslan Et al. | 2018 | Hyper-parameter Optimization | ✓ | | | | | | | | ✓ | ✓ | ✓ | | | | ✓ | | | |
| Oladele | 2021 | PSO & ACO | ✓ | ✓ | | ✓ | ✓ | ✓ | | | | ✓ | ✓ | ✓ | ✓ | ✓ | ✓ | ✓ | | |
| Rao Et al. | 2019 | ABC | ✓ | | | | | | | | | | | | | | ✓ | | | |
| Mahendru and Agarwal | 2019 | CFS Subset | ✓ | | | | | | | | | | ✓ | | ✓ | | | | | |
| Our Study | | **Proposed framework** | ✓ | ✓ | ✓ | ✓ | ✓ | ✓ | ✓ | ✓ | ✓ | ✓ | ✓ | ✓ | ✓ | ✓ | ✓ | ✓ | ✓ | ✓ |

# 3. Materials and methods

This section would like to discuss the dataset, proposed FS methods, meta-heuristics, and ML algorithms. Then, at the end of this section, we explain our framework, including the mentioned cases.

## 3.1. Wisconsin breast cancer database

The dataset used in this study was obtained from the University of California's ML division and is accessible in its database under the WDBC[1] license. This dataset includes 569 instances, each encompassing 30 real-valued input features. There are 212 malignant and 357 benign patients within this dataset. The 30 attributes are calculated from the digitalized images of a chest mass. Ten attributes with continuous values are calculated for each cell nucleus with 30 desired properties. For each picture, calculating the mean, standard error (SE), and worst or greatest (mean of the three largest values) values of these features yields 30 features. For example, the mean radius is field 1, the radius standard error is field 11, and the worst radius is field 21. Table 2 lists the features and their descriptions.

## 3.2. Proposed meta-heuristic algorithms

The ICA and the BA are surveyed in this subsection as one crucial part of the proposed framework, whose applications on disease prediction are investigated and compared.

**Table 2**
The information of features in the WDBC dataset

| Feature number | Features |
|---|---|
| 1 | Radius (mean of distances from the center to points on the perimeter) |
| 2 | Texture (standard deviation of gray-scale values) |
| 3 | Perimeter |
| 4 | Area |
| 5 | Smoothness (local variation in radius lengths) |
| 6 | Compactness (perimeter2/area − 1.0) |
| 7 | Concavity (severity of concave portions of the contour) |
| 8 | Concave points (number of concave portions of the contour) |
| 9 | Symmetry |
| 10 | Fractal dimension ("coastline approximation00 − 1) |

### 3.2.1. ICA

The ICA is an evolutionary algorithm for global optimization that provides a solution inspired by imperialistic competition. The main foundations of this algorithm are assimilation policy, colonial competition, and revolution. Like other optimization algorithms, the ICA requires the definition of different steps, whose repetition can optimize the problem (Hosseini and Al Khaled, 2014).

---

[1] Wolberg, W.H., NickStreet, W., Mangasarian, O. L., Diagnostic Wisconsin breast cancer dataset, available in, https://archive.ics.uci.edu/ml/datasets

The ICA's method may be summed up as follows:

(1) Initialization: create the $N_{pop}$ population.

(2) Create an initial empire by calculating $c_i$ for each person, sorting $c_i$ in decreasing order for all solutions, selecting $N_{imp}$ best solutions from $N_{pop}$ as imperialists, and allocating $N_{col}$ great countries to the imperialists.

(3) Assimilate colonies, carry out a revolution in certain colonies, and swap the colony and imperialist if feasible for each empire.

(4) Be victorious in imperialist competition.

(5) Destroy the empire by removing all of the nations.

(6) Continue to Step 3 if the termination requirement is not met; otherwise, cease looking.

The country cost is measured concerning the objective function: the less the cost, the better the solution. $N_{imp}$ is the most cost-effective solutions, characterized as imperialists.

The remainder of the nations has been designated as colonies. $N_{col}$ is a kind of colony that may be expressed as $N_{col} = N_{pop} - N_{imp}$ (Atashpaz-Gargari and Lucas, 2007). The first empires are formed by awarding colonies to imperialists ($k$) and conferring power ($P_k$) on the imperialists:

$$P_k = \left| \frac{Y_k}{\sum_{v=1}^{N_{imp}} Y_v} \right| \tag{1}$$

where $Y_v = max_k\{c_k\} - c_v$ denotes the normalized cost, whereas $c_k$ is the imperialist $k$'s cost. The number of initial colonies impacted by imperialist $k$ is calculated as round $\{P_k \times N_{col}\}$, where the round is a function that returns the fractional number's an adjacent integer. $H_k$ is also the collection of imperialist $k$'s colonies.

Each empire's colony alters as much as $\varepsilon$ possible along with the direction $e$ toward its imperialist throughout the assimilation process. It is important to note that $\varepsilon$ is a random number with a random probability distribution $[0, \beta \times e]$, where $\beta \in [1, 2]$. Setting $s > 1$ forces the colony to adopt an imperialist mindset. A diversion is proposed to avoid the fact that colonies are somewhat absorbed in imperialism. If there isn't a variation, the answer is more likely to be locked in a local optimum. Where $\phi$ is an arbitrary parameter, the deviation parameter $\theta$ follows a uniform probability distribution $[-\phi, \phi]$.

When it comes to the ICA's revolution, it's all about shifting certain colonies' positions owing to unanticipated differences in their qualities. A colony's traits will be altered if its language or religion is changed, leading to a shift in its location (Lei et al., 2018).

Following the assimilation and revolution processes, each colony's cost is linked to that of its imperialist, and a colony is traded with the imperialist if the colony's cost is lower. Imperialist rivalry is a major step toward an empire's ultimate might. The total cost of empire k is denoted by $TC_k$. For each empire $k$, we first calculate $TC_k$ by:

$$TC_k = c_k + \zeta \times mean\,\{Cost\,(colonies\,of\,empire\,k)\} \tag{2}$$

where $\zeta$ is a positive integer in the range of 0 to 1 that is close to 0, the normalized total cost of empire $k$ and empire $k$'s influence is then calculated as follows:

$$NTC_k = max_h\,\{TC_h\} - TC_k \tag{3}$$

$$EP_k = \left| \frac{NTC_k}{\sum_{v=1}^{N_{imp}} NTC_v} \right| \tag{4}$$

Following the definition of a vector $[EP_1 - \beta_1, EP_2 - \beta_2, \ldots, EP_{N_{imp}} - \beta_{N_{imp}}]$, the weakest colony from the poorest empire is assigned to the empire with the largest index. I denote a random integer in the range of [0, 1] associated with the uniform distribution. The ICA's pseudo-code is shown in Fig. 1.

---

Generate the population $N_{pop}$ randomly

Initialize the empire $k$

   **for** $i = 1\,to\,N_{pop}$
      Compute cost function ( fitness function) $c_i$
      Sorted the computed cost functions in descending order for the entire population
      Select $N_{imp}$ ( number of imperialist countries) out of $N_{pop}$
      Normalize the cost of each imperialist $Y_v$
      Compute the normalized power of each imperialist $P_k$
      Assign $N_{col}$ remained countries to the imperialists using Eq. (1)
   **end**

Imperialist competition process:

   **for** $j = 1\,to\,N_{imp}$
      Move colony as much as $\varepsilon$ toward direction $e$ of its relevant imperialist (assimilation)
      Calculate the costs of assimilated countries $H_k$
      Execute revolution on new colonies
      **if** ( the cost of new colony < cost of imperialist $c_k$ )
         Exchange the position of colony and imperialist
      **end**
      Pick the weakest colony from the weakest empire and allocate it to the empire with a
      higher probability of processing it using Eqs. (2) and (3)
   **end**

Removal process

      **if** (there is an imperialist with no colonies )
         Assign the imperialist as a colony to the empire that has the lowest total cost $TC_k$
      **end**

Until stopping condition is reached

---

**Fig. 1.** Pseudo-code of the proposed ICA

### 3.2.2. BA

The BA is inspired by the collective behavior of bats in their natural environment. It is based on the phenomenon of bats' treble sound reflection and the receiving of these sounds by the bat. The tool for reflecting sound and receiving its reflection in tiny bats provides them with significant hunting and crossing obstacles (Yang, 2010). The BA's steps are described as follows:

**Step 1)** Generating the primary population: Primary bats, initial parameters, and constants are determined. The initial population of bats is generated entirely randomly in the search solution space, as mentioned in Eq. (5). $X_i$ is a vector of random variables denoting the observed attributes.

$$X_i = \left(x_i^1, \dots, x_d^i, \dots, x_i^n\right) \qquad i = 1, 2, \dots, N \qquad (5)$$

**Step 2)** Calculating the objective function: At this step, the objective function, which in this paper is identical to the cost function, is determined for each bat.

**Step 3)** Updating the speed and position of bats: Each bat moves with speed $V_i^t$ and a position $X_i^t$ in the repetition of $t$ and $N$-dimension search space and moves towards the best bat ($X_*^t$). Also, each bat has a random frequency ($f_i$) in the range of $[f_{min}, f_{max}]$. Then, according to Eqs. (6) to (8), new temporary positions are generated with frequency adjustment and updating the speed of all bats. In this study, the initial velocity of each bat is set to zero, and their initial positions are selected randomly (Yang, 2010; Yang and He, 2013).

$$f_i = f_{min} + (f_{max} - f_{min})\beta \qquad (6)$$
$$V_i^t = V_i^{t-1} + (X_i^t - X_*^t)f_i \qquad (7)$$
$$X_i^t = X_i^{t-1} + V_i^t \qquad (8)$$

**Step 4)** Evaluating the chances of bats and updating their velocity and pulse emission rate: Each bat has a chance to receive a pulse from the top bat. The pulse emission rate ($R_i$) must be less than a random value ($rand$) to detect this issue. Then, the position, velocity, and $Ri$ of the bats are updated in case the new position of each bat is checked using two conditions, in which both of those conditions hold. First, the new position of each bat must be better than its previous position, and second, the loudness of the bats' voice ($A_i$) must be higher than the random value. The related equations are shown below.

$$if\ rand > R_i \qquad (9)$$
$$x_{new} = X_*^t + \varepsilon A^t \qquad (10)$$
$$if\ (\ rand < A_i\ \&\ f(x_{new}) < f(x_i)) \qquad (11)$$
$$A_i^{t+1} = \alpha A_i^t \quad,\ R_i^{t+1} = R_i^0[1 - \exp{(-\gamma t)} \qquad (12)$$

In the proposed formulas, $\beta$ represents a random number falling in the range of $[0,1]$ related to the uniform distribution in $[0, 1]$, $\varepsilon$ indicates a random number in $[-1, 1]$, and $\alpha$ and $\gamma$ are constant (Qasim and Algamal, 2020).

**Step 5)** Controlling the convergence: In convergence, the optimization stops; otherwise, the algorithm returns to Step 2, and then, Steps 2 to 5 are repeated. After completing the iterations of the BA, the position of the best bat will be the solution to the optimization problem. The BA's pseudo-code is displayed in Fig. 2.

---

$f(X),\quad X = (x_1, \dots, x_d)^T$ Objective function

$X_i$ ($i = 1, 2, \dots, n$) Initialize the bat population

Define pulse frequency $f_i$ at $X_i$

Initialize pulse rate $r_i$ and the loudness $A_i$

**While** ($t < Max\ number\ of\ iterations$)

    Produce new solutions by regulating frequency,

    and bringing up-to-date velocities and locations/solutions using Eqs. (6) to (8)

        **if** ($rand > r_i$)

            Choose a solution among the best solutions

            Generate a local solution around the best solution

        **end if**

    Create a new solution by flying randomly

        **if** ($rand < A_i$ & $f(X_i) < f(X_*)$)

            Accept the new solutions using Eq. (10)

            Increase $r_i$ and reduce $A_i$ using Eq. (12)

        **end if**

    Order the bats and find the current best $X_*$

**end while**

Post-process computational results and imagining

---

**Fig. 2.** Pseudo-code of the proposed BA

*3.3. FS methods using the ICA and BA*

This section explains the importance of using the FS method in the proposed framework. The FS chooses a minimum set of features by leaving out irrelevant features. The result of the mining task on the minimum set becomes more satisfactory than that of the complete feature set. Overall, the benefits of feature selection are as follows (Kaufmann, 1999):

- Reducing the curse of dimensionality.
- Generalizing the model.
- Speeding up the model building process.
- Enhancing the interpretability and comprehensibility of the resulting model.

The most general methods utilized in the FS are the wrapper and filter methods. The filter approach is a simple strategy by using the features are sorted based on criteria. As a whole, filter approaches are computationally inexpensive and straightforward. Meanwhile, selecting an appropriate learning algorithm can be challenging. On the other hand, under the wrapper approach, performance optimization of an ML model is used to choose the best feature subset. Wrapper approaches use a machine learning algorithm, and a metaheuristic optimization strategy is employed to get the best feature set. Besides, the classification process using the best features will be more straightforward and accurate (Qasim and Algamal, 2020). This work uses a wrapper feature selection (WFS) technique based on the BA and ICA independently and mixes them with ML algorithms to train and build classification models. The ICA and BA were selected as the metaheuristic optimization algorithms in this WFS approach because of their excellent performance (Hosseini and Al Khaled, 2014; Ahmadi et al., 2013; Lu wt al., 2020; Salma and Doreswamy, 2020).

### 3.4. ML algorithms

This section entirely describes the ML algorithms used in the presented framework. In addition, it notes that basic models of the algorithms are utilized for diagnosis.

### 3.4.1. ANN

Neural networks are interconnected processing elements known as neurons that solve a problem. One of the key advantages of ANN is that it can learn to explain complicated real-world data that is noisy, such as x-ray pictures and medical records, without explicitly doing so (Smith, 1993). An ANN comprises three layers hidden, input, and output layers. The neurons in every layer are linked with a specific weight. The weights are initialized randomly at first and then updated based on the output of the ANN such that the amount of the output is less compared to the previous iteration. The number of nodes at the input layer equals the number of features. The number of labels is represented by the number of nodes in the output layer. Trial and error are used to determine the number of nodes in the hidden layer. Furthermore, each node has a transfer function that creates output within an acceptable range (Aslan et al., 2018).

### 3.4.2. SVM

The SVM is a supervised ML algorithm employed for regression and classification Vapnik and Chervonenkis (Vapnik and Chervonenkis, 1991). Each data sample in the SVM is represented as a dot on the data scatter plot in N-dimensional space, where $N$ is the number of features in the data sample. In addition, the value for each attribute represents one of the components of the point's coordinates on the plot. Then, it classifies different and distinct data by drawing a straight line. One of the main benefits of the SVM is that, unlike empirical risk minimization-based statistical learning techniques, it tries to reduce structural risk, demonstrating a great capacity to prevent overfitting (Wang et al., 2018).

### 3.4.3. NB

The NB classification method exploits the Bayesian theorem and independence assumption between variables. It is recognized as a member of the family of probability-based classifications (Mandal, 2017). Bayesian classifiers determine the most probable class for a given case based on its feature vector. Because of the independence assumption, NB saves time. Another advantage is that NB needs a smaller training dataset than other machine learning techniques (Rish, 2001).

### 3.4.4. KNN

KNN is a learning method that is based on instances. It implies it uses whole training instances to predict output for unknown data rather than learning weights from training data. The $K$ Nearest data points are used to anticipate the new Datapoint's class or continuous value. Furthermore, since KNN is a non-parametric algorithm, the mapping function has no fixed shape (Coomans and Massart, 1982). Because there is no training period, the strategy is simple to comprehend and apply (Aslan et al., 2018).

### 3.4.5. LR

LR is a supervised probabilistic machine learning technique that predicts a category target variable. The purpose of LR is to develop the most economical and best-fitting model to explain the connection between a collection of independent (predictor or explanatory) variables and an outcome (dependent or response variable). This method is reasonably robust, adaptable, and simple to apply, and it lends itself well to interpretation. LR, unlike LDA, does not make any assumptions about the distribution of the explanatory variables (Tolles and Meurer, 2016; Pohar et al., 2004).

### 3.4.6. DT

A DT is a supervised machine learning technique that uses a directed graph to describe probable answers to a decision under specified parameters. The graph is made up of nodes that are divided into three groups: choice, chance, and terminal nodes. A chance node depicts the likelihood of particular outcomes. A decision node represents a choice that has to be taken, and a terminal node represents the decision path's conclusion (Kamiski et al., 2018). The DT algorithms provide great accuracy, stability, and readability prediction models. DTs have minimal variance because they make no assumptions about the target variable (Hamsagayathri and Sampath, 2017).

### 3.4.7. LDA

The association between a categorical variable and a group of associated factors is the focus of the LDA. It is a collection of multivariate statistical techniques for determining a linear combination of features that distinguishes or characterized two or more classes of objects or occurrences. The trained classifier of LDA finds the class with the most negligible misclassification cost to predict new data (McLachlan, 2004).

### 3.4.8. RF

The RF is a set of tree predictors in which each tree is determined by the values of a random vector collected independently and uniformly across all trees in the forest. As the number of trees in a forest becomes larger, the generalization error converges to a point (Ho, 1995; Breiman, 1996). The downside of DT classifiers is that they are prone to overfitting the training set. The RF's ensemble architecture compensates for this and enables it to generalize effectively to new data. They are also adept at dealing with massive datasets with high dimensionality (Dietterich, 2000).

### 3.4.9. AB

In machine learning, the AB is a boosting approach that is employed as an ensemble method. The weights are reassigned to each instance in this procedure, with greater weights being applied to erroneously categorized occurrences. By merging weak learners, it repeatedly corrects the faults of the weak classifier and increases accuracy. It is also resistant to overfitting (Freund and Schapire, 1997; Kégl, 2013).

### 3.5. Evaluation criteria

In this research, the following creditable criteria are exploited to evaluate and compare the efficiency of ML algorithms in diagnosing the benignity and malignancy of disease. To the best of our knowledge, true negative (TN) and true positive (TP) rates are suggested to evaluate the classifier's performance. The confusion matrix, shown in Table 3, is used to calculate TP and TN. As shown in Table 3, TP represents the number of true positives or instances that were accurately classified as benign tumors. FN stands for false negatives, which are benign tumor instances that have been misclassified as malignant; TN is for true negatives, which are accurately labeled cases in the malignant tumor. The number of false positives (FP) is the number of malignant tumor cases that were misclassified as benign.

**Table 3**
Confusion Matrix

|  | Predicted positive (benign) | Predicted negative (malignant) |
|---|---|---|
| Actual positive (benign) | True positive(TP) | False negative (FN) |
| Actual negative (malignant) | False positive (FP) | True negative (TN) |

### 3.5.1. Accuracy

The measure of accuracy represents the percentage of correct separation of data into their corresponding classes. It is well-defined as the number of correctly classified data into their related classes over the total data in the set for classification models. It is formulated by:

$$Accuracy = \frac{TP + TN}{TP + FN + FP + TN} \tag{13}$$

### 3.5.2. sensitivity

The fraction of positive cases that are appropriately classified as positive is known as sensitivity. This measure is calculated by:

$$Sensitivity\ (TPR) = \frac{TP}{TP + FN} \tag{14}$$

### 3.5.3. specificity

On the contrary, sometimes, diagnosing the negative class becomes essential to the sensitivity measure. Specificity refers to the model's ratio of negative cases correctly diagnosed as negative samples. This measure is calculated by:

$$Specificity = \frac{TN}{TN + FP} \tag{15}$$

### 3.5.4. precision

The precision measure denotes that the ratio of the data the classifier recognizes as positive is truly positive. This criterion is defined by:

$$Precision = \frac{TP}{TP + FP} \tag{16}$$

### 3.5.5. F-Score

This measure is defined as the combination of sensitivity and precision criteria and is the harmonically average of these two criteria, defined as follows:

$$F - Score = 2 \times \frac{(\text{Recall} \times \text{precision})}{(\text{Recall} + \text{precision})} \tag{17}$$

### 3.5.6. Kappa statistic

The Kappa statistic is a prominent measure for evaluating model compatibility. This statistic compares the suggested model's findings to the results provided by the stochastic classification approach. This criterion can be calculated using Eqs. (18) to (20).

$$K = \frac{[P(A) - P(E)]}{[1 - P(E)]} \tag{18}$$

$$P(A) = (TP + TN)/N \tag{19}$$

$$P(E) = [(TP + FN) \times (TP + FP) \times (TN + FN)]/N^2 \tag{20}$$

### 3.5.7. MAE

It is equal to the average of the discrepancies between the observed and projected values. This error is the average prediction error. The value for this index is defined by:

$$MAE = \frac{1}{n} \times \sum_{i=1}^{n} \left| x_{imeas} - x_{ipred} \right| \tag{21}$$

where $n$, $x_{ipred}$, and $x_{imeas}$ represent the number of measured variables, the amount of the predicted variable, and the amount of the measured variable, respectively.

### 3.5.8. RMSE

The RMSE is also a fit or objective function and is the root of the mean square error index. This index indicates the absolute error among the simulation and observation variables. The value of this index is stated by:

$$RMSE = \sqrt{\frac{1}{n} \times \sum_{i=1}^{n} \left[ \left( x_{imeas} - x_{ipred} \right)^2 \right]} \tag{22}$$

### 3.5.9. RAE

The RAE is a method for evaluating the performance of a predictive model. As shown below, it is a ratio calculated by dividing the absolute error by the variable's actual value.

$$RAE = \frac{[\sum_{i=1}^{N} (x_{ipred} - x_{imeas})^2]^{1/2}}{[\sum_{i=1}^{N} x_{imeas}^2]^{1/2}} \tag{23}$$

### 3.6. Proposed framework

This section describes the proposed framework illustrated in Fig. 3. The goal, according to it, is to improve the effectiveness of diagnostic models by combining the two novel WFS approaches based on the ICA and BA with ML algorithms for breast cancer prediction. The most significant features from the Wisconsin breast cancer dataset WDBC from the UCI machine-learning repository[1] are obtained using the WFSIC and WFSB algorithms. Furthermore, we applied nine machine learning methods that have never been used before, making our research more comprehensive in terms of assisting doctors. In reality, we use a variety of machine learning methods since each algorithm has its own set of strengths and limitations and acts differently depending on the dataset. As a result, testing with a variety of methods ensures that the best model is chosen and that variations in choosing the best features are kept to a minimum (Johnson et al., 2020).

---

[1] Wolberg, W.H., NickStreet, W., Mangasarian, O. L., Diagnostic Wisconsin breast cancer dataset, available in, https://archive.ics.uci.edu/ml/datasets

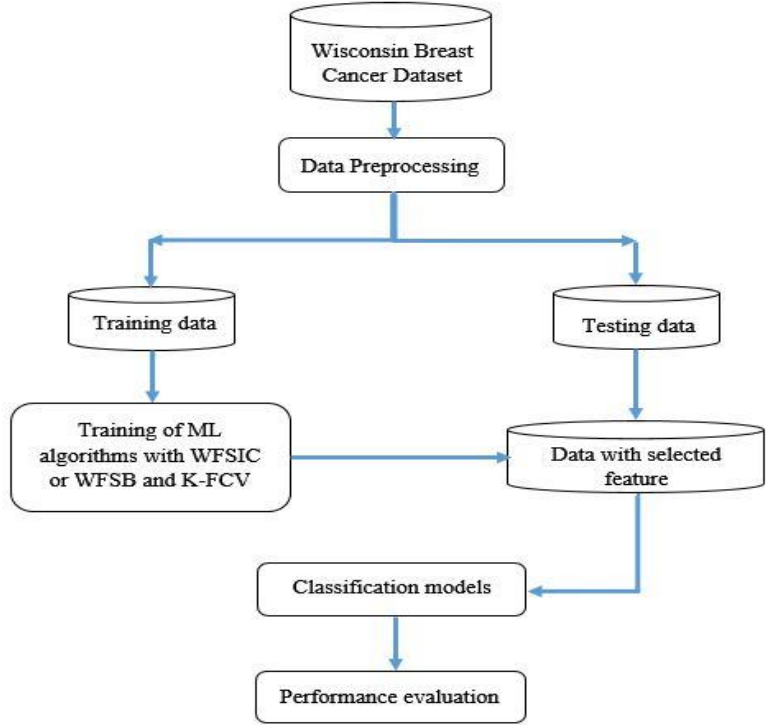

**Fig. 3.** Flowchart of the proposed framework

Preprocessing, which includes filling in missing data, eliminating outliers, splitting data, and normalizing, is done in the first step of our architecture. The data is separated into two pieces in the dividing section: training and testing data. To normalize the distribution and boost the success rate, normalization must be used. For normalizing, the feature scaling approach is utilized (Hosseini and Al Khaled, 2014). This method's formula is as follows:

$$x' = \frac{X - X_{min}}{X_{max} - X_{min}} \tag{24}$$

After that, in the second stage, for the WFSB or the WFSIC, they are individually integrated with ML algorithms along with the K-fold cross-validation (K-FCV) method and used to train them and reduce the dimensions of the clean dataset obtained from stage one. After training ML algorithms with the proposed WFS techniques and K-FCV, classification models are generated, and each ML algorithm creates the best feature set to assess their classifiers. For the WFSB, All the particles are randomly initialized using Eq. (5). After the initialization, each particle's acceleration, frequency, and position are calculated using Eqs. (6) to (8), respectively. Finally, the velocity and position of each particle are updated using Eqs. (9) to (12). This process is continued until the stopping criterion is satisfied (Qasim and Algamal, 2020). In our experiment, the end criterion is the maximum number of iterations.

For the WFSIC, all countries that are feature subsets are initialized randomly. After that, imperialists are selected based on the lowest cost of countries, and others are assigned to imperialists as colonies using Eq. (1). Then, assimilation and revolution operations are performed, and the cost of empires is calculated using Eq. (2) and compared with other colonies. If it is more than others, its position will be exchanged. Finally, the weakest colony from the weakest empire changes to one, which is more likely to process it using Eqs. (3) and (4) (Atashpaz-Gargari and Lucas, 2007; Lei et al., 2018).

Then, in the third stage, the dataset with selected features acquired from the WFSIC or WFSB integrated with ML algorithms is applied to evaluate diagnostic models generated by training in the second stage. On the one hand, Modern ML algorithms often have hyper-parameters to adjust. That is why; we deploy hyper-parameter tuning manually by trial and error for some ML algorithms. Furthermore, changing the settings of certain hyper-parameters may have little impact on the model's performance. As a result, it is suggested that the essential hyper-parameters for each method be examined (Probst et al., 2019). In ANN, for example, the ideal value for the number of hidden layers is five (Johnson et al., 2020). Furthermore, for LR and NB, we do not test any hyper-parameters. The LR features a regularisation hyper-parameter that runs an embedded FS. The NB contains the smoothing parameter alpha that controls whether or not Laplace smoothing is used during the training process. Due to the use of WFS techniques and Laplace smoothing, there is no need to tweak hyper-parameters for LR and NB. Nevertheless, we apply the K-FCV method to tune hyper-parameter $K$ for the KNN and obtain its best value at five (Tohka and van Gils, 2021).

Meanwhile, Model validation is crucial in assessing ML models' effectiveness to avoid overfitting. Because of that, in this stage, nine ML algorithms are trained with the K-FCV method to validate diagnostic models' performance. Moreover, the K-FCV method minimizes the dependency of results on the randomly chosen training samples during the estimation of a predictor to ensure that results on unobserved samples are robust.

K-FCV divides the training dataset into k-subsets of nearly similar size at random. The ML method is trained repeatedly using the $k$-1 folds and then tested on the final remaining fold, allowing each fold to be used as a test case once. The overall K-FCV (OCV) is calculated by averaging the $k$ folds' key performance indicators ($F_i$) as follows[1]:

$$OCV = \frac{1}{K} \sum_{i=1}^{k} F_i \qquad (25)$$

When $k$ is set too high, a small number of sample combinations are generated, limiting the number of alternative iterations. If $K$ is set too low, different training sets will appear throughout the folds, increasing bias. Because empirical investigations have demonstrated that models derived from four-FCV do not suffer from large variation, $k$ is chosen as four in this research.

---

[1] https://www.mathworks.com/discovery/cross-validation.html

Furthermore, four-FCV allows for a compromise between computational cost and model variance (Azar and El-Said, 2014; Olson and Delen, 2008).

The dataset in this research is split into two parts: training and testing. The training set comprises 60% of the observations in the dataset, whereas the test set contains 40% of the observations.

The ML algorithms are trained using the training dataset and four-FCV. The test set is used to determine how well the tweaked ML models perform. The 60:40 split was chosen to guarantee enough training data for the machine learning algorithms to observe more instances and train using diverse patterns. The 60:40 split also ensures enough cases in the test set to determine how well the ML models generalize to new data (Johnson et al., 2020).

In the fourth stage, results are obtained from these ML models in a situation with and without proposed approaches. Nine evaluation measures are compared, and the best model is deployed to predict breast cancer. The proposed FS algorithms (i.e., WFSIC and WFSB) are executed in MATLAB 2016b platform on an Intel Core i5 CPU with 2.40GHz, 4GB RAM, 64 bit Windows 8.1 OS. In addition, ICA parameters used in our approach are indicated in Table 4 and acquired from (Atashpaz-Gargari and Lucas 2007). BA parameters used in our approach are indicated in Table 5 and are obtained from (Yang 2010).

**Table 4**
Parameters used in the ICA

| Parameter | Explanation | Value |
|-----------|-------------|-------|
| $N_{pop}$ | Population size | 10 |
| $N_{imp}$ | Number of the most powerful countries | 5 |
| $max_{it}$ | Maximum number of iterations | 30 |
| $N_{col}$ | Number of colonies | 5 |
| $\beta$ | Assimilation coefficient | 2 |
| $\zeta$ | Colonies mean cost coefficient | 0.1 |

**Table 5**
Parameters used in the BA

| Parameter | Explanation | Value |
|-----------|-------------|-------|
| $N_{pop}$ | Population size | 10 |
| $max_{it}$ | Maximum number of iterations | 30 |
| $A$ | Loudness | 0.9 |
| $R$ | Pulse rate | 0.6 |
| $V_0$ | Initial velocity of each particle | 0 |
| $f_{min}$ | Minimum frequency | 0 |
| $f_{max}$ | Maximum frequency | 2 |

# 4. Computational results

## 4.1. Comparative analysis

### 4.1.1. ML algorithms integrated with the WFSIC method

For the WFSIC method, the performance results are demonstrated in Table 6. Concerning Accuracy and Sensitivity, all models are improved. AB with a ratio of 98.95% and LR with a ratio of 98.58% have the best execution in terms of accuracy and sensitivity, respectively. From the specificity point of view, DT, ANN, AB, and RF classifiers are increased so that AB has higher proficiency with a ratio of 99.44%. In contrast, the best specificity performance is associated with the LDA model with a ratio of 99.71% when we do not exploit the WFSIC approach. However, this model is not appropriate since its sensitivity is low and does not balance Sensitivity and Specificity. A good prediction model should make a reasonable balance between Sensitivity and Specificity. When looking at these two measures, we notice that the three models of the ANN, AB, and RF have made a good balance between Sensitivity and Specificity such that the AB is the best of them thanks to higher values. Concerning the Kappa statistic, with a ratio of 97%, the NB method has the highest compatibility against a random classifier model. According to Precision and *F*-score that is harmonically average of Sensitivity and Precision, the best implement is related to the AB with a ratio of 99.1% and 98.6%, respectively. In terms of MAE, RMSE, and RAE, the AB has revealed better performance with the lowest values.

Another conclusion that can be taken by Table 7 would be that the efficiency of all models has been reinforced given the WFSIC approach in accuracy, sensitivity, Kappa, MAE, and RMSE. Furthermore, the performance of five of nine models, including the ANN, AB, KNN, LDA, and NB, is dramatically improved.

By and large, the comparison of all evaluation criteria illustrates that the AB classifier is chosen as the best diagnostic model in combination with the WFSIC approach and is classified the dataset with 17 features. Nevertheless, suppose the performance of models based on the number of features is considered. In that case, the NB classification model has the lowest number of features among all models by using only eight features. The number of the best features resulting from each method is shown in Table 6. Note that the best results are shown in boldface.

**Table 6**
Performance results of the ML models with the WFSIC

| Models | Evaluation measures (%) | | | | | | | | | No. of selected features |
|--------|----------|-------------|-------------|-----------|---------|-------|-----|------|-----|-------------------|
| | Accuracy | Sensitivity | Specificity | Precision | *F*-Score | Kappa | MAE | RMSE | RAE | |
| RF | 98.06 | 97.64 | 98.31 | 97.2 | 97.4 | 95 | 1 | 13 | 8 | 16 |
| ANN | 98.06 | 97.64 | 98.31 | 97.2 | 97.4 | 92 | 1 | 13 | 13 | 16 |
| AB | **98.95** | 98.11 | **99.44** | **99.1** | **98.6** | 95 | **1** | **10** | **6** | 17 |
| LDA | 97.18 | 93.39 | **99.44** | 99.0 | 96.1 | 90 | 2 | 16 | 9 | 12 |
| NB | 96.30 | 94.81 | 97.19 | 95.3 | 95.0 | **97** | 3 | 19 | 19 | **8** |
| SVM | 95.43 | 94.33 | 96.07 | 93.5 | 93.9 | 91 | 4 | 21 | 12 | 9 |
| KNN | 97.53 | 96.69 | 98.03 | 96.7 | 96.7 | 93 | 2 | 15 | 9 | 18 |
| DT | 95.78 | 93.39 | 97.19 | 95.2 | 94.3 | 90 | 4 | 20 | 12 | 21 |
| LR | 97.01 | **98.58** | 96.07 | 93.7 | 96.1 | 95 | 2 | 17 | 10 | 14 |

**Table 7**

Performance results of ML models without the proposed approaches

| Models | Evaluation measures (%) | | | | | | | | |
|---|---|---|---|---|---|---|---|---|---|
| | Accuracy | Sensitivity | Specificity | Precision | *F*-Score | Kappa | MAE | RMSE | RAE |
| RF | **97.89** | **97.16** | 97.16 | 97.2 | **97.2** | **92** | **2** | **14** | **8** |
| ANN | 95.60 | 95.60 | 96.35 | 93.9 | 94.1 | 90 | 4 | 20 | 12 |
| AB | 95.95 | 95.75 | 96.07 | 93.5 | 94.6 | 91 | 4 | 20 | 11 |
| LDA | 95.60 | 88.67 | **99.71** | **99.5** | 93.8 | 90 | 4 | 20 | 12 |
| NB | 94.20 | 89.15 | 97.19 | 95.0 | 92.0 | 87 | 5 | 24 | 14 |
| SVM | 94.55 | 91.98 | 96.07 | 93.3 | 92.6 | 88 | 5 | 23 | 13 |
| KNN | 93.14 | 84.90 | 99.01 | 96.3 | 90.0 | 84 | 6 | 26 | 15 |
| DT | 94.37 | 91.50 | 96.07 | 93.3 | 92.4 | 87 | 5 | 23 | 13 |
| LR | 96.13 | 95.28 | 96.63 | 94.4 | 94.8 | 91 | 3 | 19 | 11 |

**Table 8**

Features selected by training ML algorithms with WFSIC and K-FCV

| | RF | ANN | LDA | SVM | KNN | AB | DT | LR | NB |
|---|---|---|---|---|---|---|---|---|---|
| 1. Radius_Mean | | + | | | | | + | + | |
| 2. Texture_Mean | + | | | | | + | | | + |
| 3. Perimeter_Mean | + | | | + | + | + | + | | |
| 4. Area_Mean | + | + | | | | + | + | + | |
| 5. Smoothness_Mean | | | | + | + | + | + | | + |
| 6. Compactness_Mean | + | | | | | + | + | + | |
| 7. Concavity_Mean | | + | | + | | | + | + | |
| 8. Concave points_Mean | + | | + | | + | + | + | | |
| 9. Symmetry_Mean | | | | + | + | + | + | | + |
| 10. Fractal dimension_Mean | + | + | | + | + | + | + | | |
| 11. Radius_SE | | + | | | + | | + | | |
| 12. Texture_SE | | | | + | | | | + | |
| 13. Perimeter_SE | | | | + | + | + | | + | |
| 14. Area_SE | + | + | + | | + | + | + | + | |
| 15. Smoothness_SE | + | | + | | + | + | + | | + |
| 16. Compactness_SE | + | + | + | | | | + | | |
| 17. Concavity_SE | + | + | + | | + | | + | + | + |
| 18. Concave points_SE | | + | + | | + | + | + | | |
| 19. Symmetry_SE | | | + | | | + | + | | |
| 20. Fractal dimension_SE | | + | | + | + | + | + | | |
| 21. Radius_Worst | + | + | + | | | + | + | + | |
| 22. Texture_Worst | + | + | + | + | + | | | + | + |
| 23. Perimeter_Worst | | + | + | + | | | | | |
| 24. Area_Worst | + | | + | | | | | + | + |
| 25. Smoothness_Worst | | + | + | | + | + | | | |
| 26. Compactness_Worst | + | + | + | | + | + | + | + | |
| 27. Concavity_Worst | | | + | | + | | + | | |
| 28. Concave points_Worst | + | | + | | | | + | + | + |
| 29. Symmetry_Worst | + | + | + | + | | + | + | | |
| 30. Fractal dimension_Worst | | | | | | | + | | |

The WFSIC technique identifies the best features for each ML model, as shown in Table 8. The findings show that various machine learning models create diverse feature sets. Nonetheless, among the feature sets found, several models have a lot in common. Seven ML methods, for example, choose Area SE, Concavity SE, Texture Worst, and Compactness Worst, suggesting that these features may be important in breast cancer prediction. Six methods choose Concave points Mean, Fractal dimension Mean, Smoothness SE, Radius Worst, and Symmetry Worst. These features may also aid doctors in detecting breast cancer in its early stages. It also saves money since suppliers no longer have to develop additional features.

### 4.1.2. ML algorithms integrated with WFSB method

The performance results in Table 9 indicate the combination between ML algorithms and the WFSB. From the Accuracy, Sensitivity, and specificity point of view, the RF classifier with 99.12%, 98.58%, and 99.44% ratios, respectively, outperformed all models, with only ten features. Furthermore, this model achieves better accomplishment thanks to the lowest values in the MAE, RMSE, and RAE criteria. According to Kappa statistics, with a ratio of 98%, it has the highest compatibility against a random classifier. The best execution was related to the RF classifier with 99.1% and 98.8% ratios concerning precision and *F*-score.

**Table 9**
Performance results of ML models with the WFSB

| Method | Evaluation measures (%) | | | | | | | | | No. of Selected Features |
| | Accuracy | Sensitivity | Specificity | Precision | F-Score | Kappa | MAE | RMSE | RAE | |
| --- | --- | --- | --- | --- | --- | --- | --- | --- | --- | --- |
| RF | **99.12** | **98.58** | **99.43** | **99.1** | **98.8** | **98** | **1** | **9** | **6** | 10 |
| ANN | 98.24 | 98.11 | 98.31 | 97.2 | 97.7 | 96 | 2 | 13 | 8 | 17 |
| AB | 97.71 | 96.69 | 98.31 | 97.2 | 96.9 | 95 | 2 | 15 | 9 | 17 |
| LDA | 96.66 | 92.45 | 99.15 | 98.5 | 95.4 | 92 | 3 | 18 | 11 | 11 |
| NB | 95.78 | 95.75 | 95.79 | 93.1 | 94.4 | 91 | 4 | 20 | 12 | 8 |
| SVM | 97.36 | 97.64 | 97.19 | 95.4 | 96.5 | 94 | 2 | 16 | 9 | 18 |
| KNN | 94.90 | 89.15 | 98.31 | 96.9 | 92.9 | 89 | 5 | 23 | 13 | 9 |
| DT | 95.25 | 90.09 | 98.32 | 97.0 | 93.4 | 90 | 5 | 22 | 12 | **3** |
| LR | 97.53 | 97.64 | 97.47 | 95.8 | 96.7 | 94 | 2 | 16 | 9 | 14 |

As mentioned earlier, due to balancing between Sensitivity and Specificity, five models of the SVM, NB, LR, RF, and ANN classifiers are desirable to diagnose more reliable, and the RF is superior in a balancing state since it has higher values in these two criteria. Another result we want to mention can be revealed in Table 7. However, the proficiency of all models is reinforced given the WFSB method. The two models of NB and AB have experienced a remarkable improvement. Meanwhile, the highest performance enhancement by the WFSB compared to the method without FS is concerned with the NB model, which achieved this result using eight features.

Overall, by analyzing the obtained results, we conclude that the RF classifier acquires the best performance regarding all criteria combined with the WFSB approach by selecting ten features. In another finding, if the performance of models based on the number of features is considered, the DT classification model has the lowest number of features by using only three features with an accuracy rate of 95.78%. Nevertheless, it is not an appropriate model to predict breast cancer since its sensitivity is low and does not balance Sensitivity and Specificity.

As demonstrated in Table 10, the best features for each ML model are identified for the WFSB approach. Like the WFSIC, there are more critical features that have been selected by most ML models in the WFSB approach. For example, Texture_Worst is the feature that has been selected by seven models. Moreover, Six models have chosen Perimeter_SE and Radius_Worst. Also, Compactness_Mean, Concavity_Mean, Smoothness_Worst, Concavity_Worst, and Symmetry_Worst have been chosen by five models, which means that these features are more critical than others.

**Table 10**
Features selected by training ML algorithms with WFSB and K-FCV

|  | RF | ANN | LDA | SVM | KNN | AB | DT | LR | NB |
|---|---|---|---|---|---|---|---|---|---|
| 1. Radius_Mean |  | + |  |  | + |  | + |  | + |
| 2. Texture_Mean |  |  |  |  |  |  |  |  |  |
| 3. Perimeter_Mean |  | + |  | + |  | + |  | + |  |
| 4. Area_Mean |  |  |  |  |  |  |  |  |  |
| 5. Smoothness_Mean |  |  | + | + |  | + |  |  |  |
| 6. Compactness_Mean | + | + | + |  |  | + |  | + |  |
| 7. Concavity_Mean |  | + | + | + |  | + |  | + |  |
| 8. Concave points_Mean |  | + |  |  |  | + |  | + |  |
| 9. Symmetry_Mean |  |  |  | + | + |  |  |  |  |
| 10. Fractal dimension_Mean |  |  |  | + |  |  |  |  |  |
| 11. Radius_SE |  | + |  | + | + | + |  |  |  |
| 12. Texture_SE | + |  | + |  | + | + |  |  | + |
| 13. Perimeter_SE | + | + | + |  |  | + |  | + | + |
| 14. Area_SE |  | + | + |  |  |  |  | + |  |
| 15. Smoothness_SE |  | + |  | + | + | + |  |  |  |
| 16. Compactness_SE |  |  |  | + |  | + |  |  |  |
| 17. Concavity_SE |  | + | + | + |  | + |  |  |  |
| 18. Concave points_SE | + |  |  | + |  | + |  |  |  |
| 19. Symmetry_SE |  |  |  |  |  |  |  | + |  |
| 20. Fractal dimension_SE | + |  |  | + |  |  |  | + | + |
| 21. Radius_Worst | + |  | + | + | + |  |  | + | + |
| 22. Texture_Worst | + | + | + | + |  | + |  | + | + |
| 23. Perimeter_Worst |  |  |  | + |  | + |  |  |  |
| 24. Area_Worst | + |  | + | + |  |  | + |  |  |
| 25. Smoothness_Worst | + | + | + | + |  |  |  | + |  |
| 26. Compactness_Worst | + | + |  |  | + | + |  | + |  |
| 27. Concavity_Worst |  | + |  | + |  | + |  | + | + |
| 28. Concave points_Worst |  | + |  |  |  |  |  |  | + |
| 29. Symmetry_Worst |  | + |  | + | + | + |  | + |  |
| 30. Fractal dimension_Worst |  | + |  |  | + |  | + |  |  |

*4.2. Comparison between the performances of WFSIC and WFSB*

Several issues can be stated when comparing the results obtained from the WFSIC and WFSB and investigating Tables 6, 7, 9, and 11. First, it should be noted that all classification models had performance advancements in these techniques. However, suppose each approach is evaluated and compared separately. In that case, five models have an extraordinary development in the WFSIC versus without it since they have significantly improved in the minimum of four criteria or their primary (i.e., accuracy, sensitivity). For example, the highest rate of improvement is related to the KNN, which has experienced a growth of about 16 percent in accuracy and sensitivity altogether. In return, with the WFSB method, two of the nine existing models have experienced extra performance reinforcement, indicating that selecting feature is more effective with the WFSIC than the WFSB. Second, about the lowest number of features chosen, WFSB integrated with the DT algorithm with selecting three features acts superiorly compared to WFSIC combined with the NB with choosing eight features. Third, in terms of overall efficiency, WFSB integrated with the RF classifier provided superior performance by recording better evaluation criteria and selecting fewer features than the WFSIC combined with the AB classifier.

**Table 11**
Comparing the performance of the proposed framework between the ICA and the BA.

| Scale | Proposed approaches | |
| --- | --- | --- |
| | **WFSIC** | **WFSB** |
| Best model | AB | RF |
| Accuracy of the best model | 98.95% | 99.12% |
| Number of the selected features by the best model | 17 | 10 |
| Least number of selected features among all models | 8 | 3 |
| Number of notable improvements | 5 models | 2 models |

Table 12 compares the results of previous research in the literature with the approaches proposed in this study. The evaluation criteria and ML models utilized in each study are stated in this table. According to this table, there are several occasions why our research is more authentic and reliable than the other ones. One of the main reasons would be that most papers have utilized just a few very common criteria such as accuracy, sensitivity, and specificity that could not be comprehensive to make more precise decisions for physicians. We have used nine criteria and nine ML algorithms together that make our work more reliable. Besides, the current research has outperformed all evaluation measures among all studies that have used FS methods cited in Table 12 and most previous papers.

Another rationale is that the highest balancing between Sensitivity and Specificity derives from our paper, while there are vast differences between these two measures among other studies. Hence, our classifiers are more reliable to employ in real clinical predictions. As we mentioned earlier, this study exploits nine possible ML models for classification problems that have never been used before and utilize more evaluation measures than most current studies. Therefore, this paper could potentially be considered one of the most validating and reliable works performed.

**Table 12**
Comparing the performance of ML models with the proposed FS methods against existing FS methods

| Authors | Year Published | Methods | ML model | Dataset | Evaluation measures (%) | | | | | | | | |
|---|---|---|---|---|---|---|---|---|---|---|---|---|---|
| | | | | | Accuracy | Sensitivity | Specificity | Precision | *F*-Score | Kappa | MAE | RMSE | RAE |
| Sakri et al. | 2018 | PSO | NB, KNN, DT | WPBC | 81.3 | 93.6 | 63.2 | 88.3 | 87.7 | 49 | 13 | 38 | 64 |
| Oladele et al. | 2021 | PSO & ACO | SVM, KNN, NB, LR, RF | WDBC | 97.13 | 97.5 | | 97.2 | 97.1 | 93.7 | | | |
| Hamsagayathri, and Sampath | 2017 | WEKA Filter | LR, NB, RF | WBC | 98.95 | 74.4 | 48.11 | | | 17 | 37 | 46 | 89 |
| Mojrian et al. | 2020 | ELM-RBF | ANN | WDBC | 95.69 | 97.56 | 94.39 | **99.1** | 98.3 | | | 15 | |
| Sangaiah and Kumar | 2019 | RF-EGA | SVM, NB, KNN | WBC | 85.89 | 76 | 88 | 81 | 82.5 | | | | |
| Salma and Doreswamy | 2020 | Hybrid BATGSA | ANN | WDBC | 92.16 | 94.3 | 89.36 | 84 | | | | | |
| Salehi Et al. | 2020 | MLP stacked generalization | ANN | SEER | 84.32 | 82.6 | 88.6 | | | | | | |
| Mohammad Ubaidullah Bokari | 2017 | Gini & Entropy | DT | WDBC, WBC | 90.5 | 92.2 | 90.4 | | | | | | |
| Fan and Chaovalitwongse | 2010 | LAD | SVM | WDBC | 96.01 | | | | | | | | |
| Mahendru and Agarwal | 2019 | Cfs subset | KNN, SVM | WDBC | 97.3 | | | | | | | | |
| Liu et al. | 2019 | IGSAGAW | KNN, SVM | WDBC, WBC | 95.8 | | | | | | | | |
| Rao et al. | 2019 | ABC & GBDT | DT | WDBC | 97.18 | | | | | | | | |
| **This study** | | WFSB | RF, ANN, AB, LDA, NB | WDBC | **99.12** | **98.58** | 99.43 | **99.1** | **98.8** | **98** | **1** | 9 | 6 |
| | | WFSIC | SVM, KNN, DT, LR | | 98.95 | 98.11 | **99.44** | **99.1** | 98.6 | 97 | **1** | 10 | **6** |

## 5. Discussion and implications

Despite many advances in the healthcare and health sciences, breast cancer continues to be the main cause of mortality among women (O'Brien et al., 2007). Moreover, this disease is curable at early stages, providing the correct and timely diagnosis. Knowing these, we propose a novel framework with a comprehensive analysis to enhance treatment outcomes for breast cancer prediction with an accuracy of over 99%, which is higher than most other medical techniques. In our framework, the WFS methods based on the ICA and the BA separately and their combination with ML algorithms are applied.

Although accuracy and sensitivity might not be enough for reliable diagnosis to manage high-dimensional data sets, if specialists consider them as decision criteria, we can recommend the RF classifier integrated with WFSB. This model attains 99.12% and 98.58% accuracy and sensitivity, respectively. Besides, a valid diagnostic model balances Specificity and Sensitivity, which means the difference between them in a balanced model is low (Dziak et al., 2020). In this respect, we present the RF combined with the WFSB and the AB combined with the WFSIC that are the most balanced models than all studies indicated in Table 12 such as (Mojrian et al., 2020; Sakri et al. 2018; Salma and Doreswamy, 2020; Sangaiah and Kumar, 2019), since the difference between the two criteria of Sensitivity and Specificity is approximately one percent.

Nevertheless, if Sensitivity and Specificity are nearly the same levels, medics can exert Kappa statistics to ensure their decisions. Concerning Kappa, we indicate that the RF classifier with a ratio of 98% has the highest compatibility versus a random model, which means it is less misleading than using accuracy as a metric and is superior to previous works (Sakri et al., 2018; Hamsagayathri et al., 2017; Oladele et al., 2021)

In addition to the former criteria, for the precision and F-score, our paper recommends applying the RF model combined with the WFSB with the best performance regarding the precision of 99.1%, F-score of 98.8%, and other former criteria. Finally, we simultaneously propound error measures along with other measures for the most sustainable and trustworthy breast cancer prediction, including MAE, RMSE, and RAE with a ratio of 1%, 9%, and 6%, respectively employing the RF model integrated with the WFSB approach that are the lowest values than the current studies (Mojrian et al., 2020; Sakri et al., 2018; Hamsagayathri and Sampath, 2017). The physicians could consider error measures with the other proposed criteria in this study at the same time to make more reliable decisions. When comparing our approaches, we should assert that the WFSIC boosted the efficiency of five ML models remarkably and outperformed the WFSB regarding the number of models that had notable progress. Another proposed approach (i.e., the WFSB) had the highest scores concerning nearly all evaluation measures with fewer selected features than the WFSIC, as illustrated in Tables 6 and 9. The best features chosen by individually integrating WFSIC and WFSB with ML algorithms had in common. These features include Area_SE, Concavity_SE, Texture_Worst, Compactness_Worst, Concave points_Mean, Fractal dimension_Mean, Smoothness_SE, Radius_Worst, Symmetry_Worst, Perimeter_SE, Compactness_Mean, Concavity_Mean, Smoothness_Worst, and Concavity_Worst.

On the whole, it can be decided from the observations and analysis of the obtained results that the WFSB is the best of our approaches, which outperforms the WFSIC and other ML models by selecting only 33% (i.e., 10 features) of the dataset's features, which prompts clinical diagnosis cost and time to diminish. Furthermore, because each ML algorithm has merits and demerits, using different models with different traits makes our best model robust. Consequently, our paper could help the physicians make a more meticulous decision with less patient wait time, medical cost, and higher authenticity than most previous works. To endorse our approaches, we achieved an accuracy of over 99% using nine ML algorithms along with nine evaluation criteria together.

In terms of the research's key implications for the current literature, this publication proposed a new paradigm for breast cancer prediction. Moreover, it is essential to note that the works illustrated in Table 12, have exerted the dataset that has been used in this study and are considered as benchmarks. It was observed that the best accuracy that is related to Mahendru and Agarwal (2019) yielded 97.3%, the best sensitivity obtained was 97.56% from (Mojrian et al., 2020), the best specificity was 94.39% that is acquired by (Mojrian et al., 2020), the best precision resulted by (Mojrian et al., 2020) was 99.1%, the best yielded F-score was 98.3% from (Mojrian et al., 2020), the best Kappa obtained was 93.7% from (Oladele et al., 2021). The best RMSE acquired was 15% from (Mojrian et al., 2020). The performance metrics obtained from these investigations are lower than those obtained from the RF and AB models proposed in this study. Additionally, these papers consider fewer ML models than those assessed in our framework. Consequently, we believe that our framework has produced better outcomes and that our framework accurately and thoroughly forecasts breast cancer based on a set of medical features.

In terms of practical consequences, this work has created data mining methods that specialists may use to forecast breast cancer more accurately and distinguish the best breast cancer traits and their numbers. To this end, oncology-focused healthcare practitioners may use the proposed framework to respond more quickly and make more reliable treatment choices, enhancing treatment results and saving costs. As a result, the suggested machine learning-based tools will aid in the advancement of healthcare analytics for breast cancer prediction.

## 6. Conclusions and future work

The implications of combining WFS approaches based on the ICA and BA with ML algorithms to improve treatment outcomes with better reliability for breast cancer prediction were investigated in this research. It was demonstrated that the performance of all introduced ML algorithms could be improved by using our proposed approaches (i.e., the WFSIC and WFSB). The WFSB outperformed the WFSIC regarding nearly all evaluation criteria in our framework. In all evaluation measures, the RF classifier integrated with the WFSB achieved higher scores than other ML models combined with the WFSB or WFSIC. Moreover, the RF model has had better performance than most previous studies. The AB classifier was the best diagnostic model combined with the WFSIC approach with 17 selected features. Besides, regarding the number of models with significant performance improvement, the WFSIC with five ML models accomplished better than the WFSB with two models. For instance, the KNN underwent an approximately 16% increase in accuracy and sensitivity altogether. Furthermore the results acquired by the proposed FS approaches reveal that the nine features, including the standard error of area, concavity, smoothness, perimeter, the worst of texture, compactness, radius, symmetry and the mean of concave points, fractal dimension, compactness, concavity rank among the superlative critical features and could considerably influence the efficiency of breast cancer prediction.

The RF integrated with the WFSB and the AB integrated with the WFSIC that were the most balanced models, since the difference between the two criteria of sensitivity and specificity is approximately one percent. This illustrared that these two models were considered as better diagnostic models for breast cancer prediction.

For the number of selected features, our best model chose only 10 (i.e., 33%) features, and the DT integrated with the WFSB selects only 3 (i.e., 10%) features for breast cancer prediction. In addition, the contributions of this study reflected the superiority of the proposed techniques compared to most previous studies in terms of model performance. For example, the RF model combined with the WFSB achieved over 99% concerning the accuracy, specificity, and precision and attained over 98% regarding sensitivity, F-score, and Kappa. Hence, healthcare operators could exploit the WFSIC and WFSB to predict breast cancer in less time and costs at early stages with higher efficiency. Likewise, our research can be applied in natural clinical diagnostic systems and therapy to assist clinical physicians in making more reliable decisions thanks to using more ML models than other works, making our study more comprehensive.

This paper's limitation could be that our proposed framework was executed using only one breast cancer dataset. Among the constructive suggestions for further expanding this research, one can mention using other breast cancer datasets. Also, this study has far-reaching ramifications since it can be used to treat and investigate various forms of cancer such as lung, colorectal, and prostate cancers. Another offer is that our framework can be used for other FS methods (e.g., relief ranking and information gain) and other metaheuristic algorithms to present more comparative analysis. Moreover, we try to work on the KNN and the NB, which have attained immense boosting by our approaches.

## Data and code availability

All codes (included classification algorithms and feature selection algorithms) and datasets publicly exist at: https://github.com/kamyab24/Machine-learning-Algorithms.git

## Conflict of interest statement

The authors declared no potential conflicts of interest for the research, authorship, and publication of this article.